\pdfoutput=1

\documentclass[11pt]{article}

\usepackage{acl}

\usepackage{times}
\usepackage{latexsym}

\usepackage[T1]{fontenc}

\usepackage[utf8]{inputenc}

\usepackage{microtype}

\usepackage{inconsolata}

\usepackage{graphicx}

\title{Comparing Trajectory and Vision Modalities for Verb Representation}

\author{
    Dylan Ebert\\
    Brown University\\
    \texttt{dylan\_ebert@brown.edu}
    \\\And
    Chen Sun\\
    Brown University\\
    \texttt{chensun@brown.edu}
    \\\AND
    Ellie Pavlick\\
    Brown University\\
    \texttt{ellie\_pavlick@brown.edu}
}

\begin{document}
\maketitle
\begin{abstract}
Three-dimensional trajectories, or the 3D position and rotation of objects over time, have been shown to encode key aspects of verb semantics (e.g., the meanings of \textit{roll} vs.\ \textit{slide}). However, most multimodal models in NLP use 2D images as representations of the world. Given the importance of 3D space in formal models of verb semantics, we expect that these 2D images would result in impoverished representations that fail to capture nuanced differences in meaning. This paper tests this hypothesis directly in controlled experiments. We train self-supervised image and trajectory encoders, and then evaluate them on the extent to which each learns to differentiate  verb concepts. Contrary to our initial expectations, we find that 2D visual modalities perform similarly well to 3D trajectories. While further work should be conducted on this question, our initial findings challenge the conventional wisdom that richer environment representations necessarily translate into better representation learning for language. 

\end{abstract}

\begin{figure*}[!ht]
    \centering
    \includegraphics[width=\textwidth]{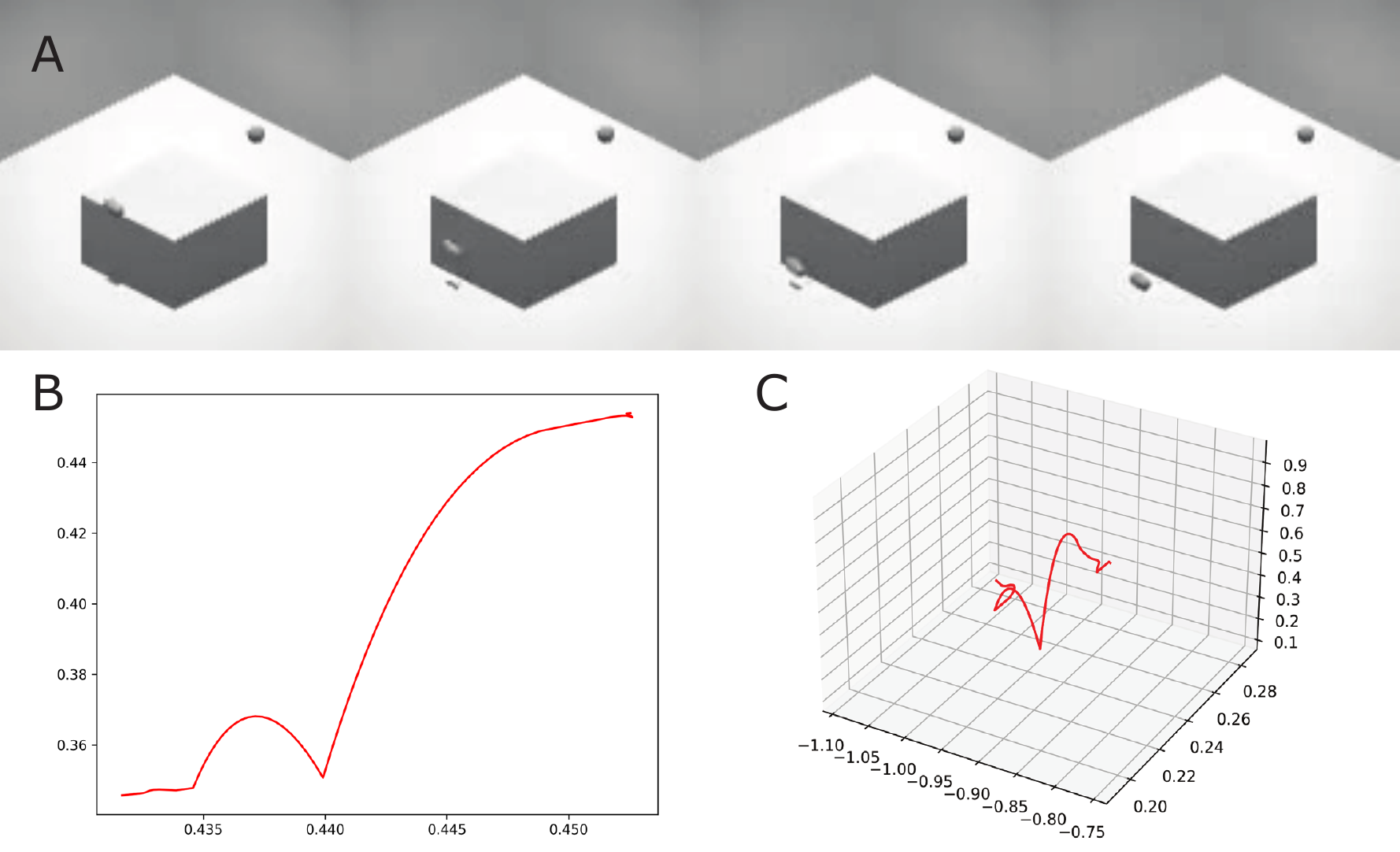}
    \caption{The \textit{Simulated Spatial Dataset} consists of procedurally generated motion data of a virtual agent interacting with an object. A) shows the camera view as the object (in red) as it falls off the counter. B) shows the corresponding 2D trajectory data, while C) shows the corresponding 3D trajectory data. These correspond to the types of features we compare.}
    \label{fig:simulated}
\end{figure*}

\section{Introduction}

Understanding and representing the meaning of verbs has long been a challenge for natural language processing. In formal semantics, verbs prove especially nuanced because of the rich physical and temporal context required to differentiate subtle syntactic and semantic differences \cite{pinker2003language}. For practical applications, verb understanding, arguably more so than nouns (i.e., object recognition), is a major bottleneck preventing AI agents from interacting effectively with humans in realistic environments \cite{shridhar-etal-2020-end,bisk-etal-2020-experience}. 

To date, most work on multimodal language understanding has exploited datasets which pair 2D visual data--e.g., images \cite{deng2009imagenet} or videos \cite{sun2019videobert}--with natural language descriptions. However, prior work has shown that such data alone is an insufficient representation for fully capturing verb semantics \cite{Yatskar_2016_CVPR}, failing to isolate verb meaning from context in which it occurs. More recently, a case has been made that 3D embodied data is the more appropriate choice for learning and grounding verb meanings \cite{ebert-etal-2022-trajectories}. Such arguments are appealing intuitively (it is difficult to capture the meaning of a word like \textit{``tumble''} without appealing to the notion of 3D space) and supported by theoretical work from formal semantics \cite{pustejovsky2016generating}. In addition, with the increasing availability of simulated environments \cite{puig2018virtualhome,juliani2018unity,gan2020threedworld}, it is becoming increasingly feasible to train models directly on 3D context. Almost certainly, we are within years of a large language model trained in a 3D interactive environment.

Despite the interest in realistic, embodied environments, there have yet to be any controlled studies comparing these richer, more cognitively-plausible learning environments to the current status quo (i.e., 2D images) in terms of their usefulness for language representation learning. There are many reasons why such controlled comparisons are challenging: 2D image data is available at a much larger scale than 3D data, the types of language used in each data set differs dramatically (as a function of the applications for which the datasets are designed), and naturalistic data is often full of confounds (e.g., a kitchen scene makes the verb \textit{``chop''} extremely likely) which make it hard to isolate the quality of the lexical semantic representations themselves. In this paper, we address these challenges by running an apples-to-apples comparison using self-supervised models trained in a stripped-down simulated data provided by \citet{ebert-etal-2022-trajectories}. This data contains videos and 3D environment recordings of abstract objects in motion, and has been labeled post-hoc for 24 individual verbs at a low level of granularity. As such, it provides an ideal test bed in which to examine the relative strengths of 2D vs.\ 3D representations for language representation learning. We find, perhaps surprisingly, that there is no clear advantage to the 3D representations compared the 2D ones in this setting. Our results do not close the book on this question: there is of course always the possibility that the picture changes with increased scale or increasingly complex verb types. However, they represent an important early step in investigating the potential and the challenges of language learning in embodied environments, and serve as an invitation to reexamine some basic assumptions about lexical semantic representations.

\section{Experimental Design}

\subsection{Dataset}

We use the Simulated Spatial Dataset from \citealt{ebert-etal-2022-trajectories} for our experiments, which contains 3000 hours of procedurally generated data of a virtual agent interacting with the object. The data is annotated with 24 verb labels, where crowdworkers label 1.5s clips with a yes/no label indicating whether each verb occurs in the clip. In total, there are 2400 annotations, with 100 annotations per verb.

This dataset was chosen for several reasons. First, it provides a relative large amount of partially-labeled trajectory data, which allows us to train our self-supervised encoders. Second, it is designed to elicit a wide variety of trajectory data. Finally, it allows greater control over experimental conditions, allowing us to test our hypothesis about the representation of verb semantics directly.

\begin{table*}[!htp]
    \centering
    \begin{tabular}{|l|c|c|}
        \hline
        Model & mAP (\% micro) & mAP (\% macro) \\
        \hline
        Random & 39.44 $\pm 1.87$ & 41.19 $\pm 1.53$ \\
        3D Trajectory & 83.38 $\pm 1.27$ & 69.95 $\pm 1.54$ \\
        2D Trajectory & 82.99 $\pm 3.97$ & 69.94 $\pm 3.09$ \\
        2D Image & 81.02 $\pm 2.74$ & 68.18 $\pm 2.95$ \\
        2D Image + 2D Trajectory & 82.38 $\pm 1.22$ & 68.80 $\pm 1.42$ \\
        2D Image + 3D Trajectory & \textbf{84.06 $\pm 1.02$} & \textbf{71.72 $\pm 1.13$} \\
        \hline
    \end{tabular}
    \caption{Mean Average Precision (mAP) scores for each model on the verb classification task, reported with both micro and macro averaging. 95\% confidence intervals are reported beside each condition.}
    \label{tab:results}
\end{table*}

\subsection{Models}

To evaluate the effectiveness of each modality for verb representation, we follow two steps:

1. Train a self-supervised LSTM encoder using a time-series prediction task, based on models from prior work \cite{ebert-etal-2022-trajectories}. The encoder is a simple feed-forward model followed by an LSTM. The input is a 90x$d$ matrix of time-series data, corresponding to a 90-frame (1.5s) clip, with $d$ dimensions depending on the modality. During self-supervised pretraining, the LSTM is unrolled 60 timesteps (1s), the outputs of which are trained to approximate true future frames using a discounted mean-squared-error (MSE) loss. This loss is discounted according to how far the prediction is in the future. Hyperparameters such as batch size, learning rate, discount factor, and hidden width were tuned using a grid search on the performance of each modality on the development data. This self-supervised pretraining is performed using the 2400-hour training subset of the Simulated Spatial Dataset.

2. Fine-tune and evaluate the encoder on a supervised verb classification task. Once the self-supervised encoder is trained, it is fine-tuned on a supervised verb classification task, using the crowdsourced annotations on the Simulated Spatial Dataset. The fine-tuning process is conducted using cross-entropy loss, and evaluated using the mean Average Precision score (mAP) on the test data.

\subsection{Features}

All of our experiments use the general self-supervised training and architecture described above, varying only the input features. The features we consider are described below.

\paragraph{3D Trajectory.} This approach uses 10-dimensional 3D trajectory data as input. That is, the 3D euclidean XYZ position of the hand and object, and quaternion XYZW rotation of the object.

\paragraph{2D Image.} This approach uses 2048-dimensional Inception-v3 \cite{szegedy2016rethinking} embeddings trained on ImageNet \cite{deng2009imagenet} as input. We also evaluated a convolutional encoder trained on our raw image data, but found that it fails to encode temporal relationships in the Simulated Spatial Dataset. As a result, we only include results from Inception in the main body of this paper. More details on the convolutional experiments can be found in Appendix \ref{app:convolutional}.

\paragraph{2D Trajectory.} This approach uses 4-dimensional 2D trajectory data as input. Specifically, this is the 2D euclidean XY position of the hand and object. The purpose of this experiment is to disentangle the real performance of the 2D image-based approach from the theoretical potential of a perfect 2D object-detection model which simply traces the trajectory of the object in 2D space.

\paragraph{2D Image + 2D Trajectory.} This approach combines the 2D image and 2D trajectory modalities, encoding each together into a shared embedding space. This approximates a perfect object detection model in conjunction with additional information that may be inferred from the visual modality.

\paragraph{2D Image + 3D Trajectory.} This approach combines the 2D image and 3D trajectory modalities, which points toward the potential for future work that may involve combining modalities.

\section{Results}\label{sec:results}

\subsection{Main Findings}

Table \ref{tab:results} shows the Mean Average Precision (mAP) scores for each model on the verb classification task. These results suggest that while all modalities perform significantly above random, there is little difference in the performance of the modalities. While the 2D Image + 3D Trajectory model does outperform other models on the verb classification task, with a mAP score of 84.06 at the micro level and 71.72 at the macro level, the difference in performance is not significant, with its 95\% confidence interval overlapping with other approaches.

\subsection{Analysis}

Table \ref{tab:verbs} shows the mAP of 3D Trajectory, 2D Trajectory, and 2D Image modalities, broken down by verbs \textit{fall} and \textit{roll}, the only verbs which exhibit a significant difference by modality. The full breakdown by verb can be found in Appendix \ref{app:full-results}. The difference for \textit{fall} can be explained by failure cases where the image-based model fails to encode verb meaning when the object becomes obscured or has very low contrast with the background. \textit{Roll}, on the other hand, is an interesting case where the image-based model outperforms trajectory-based models. In qualitative analysis, this often involves instances where a round object with low friction slides, but doesn't actually \textit{roll}, per se. We can not determine whether this is due to annotator error or conceptual differences in the meaning of \textit{roll}, but in either case we believe that this highlights the challenges of verb learning. 

\begin{table}[!htp]
    \centering
    \begin{tabular}{|l|c|}
        \multicolumn{2}{c}{Fall} \\
        \hline
        Model & mAP (\%) \\
        \hline
        Random & 26.78 $\pm 4.63$ \\
        3D Trajectory & 96.56 $\pm 2.63$ \\
        2D Trajectory & 95.22 $\pm 3.55$ \\
        2D Image & 87.67 $\pm 4.33$ \\
        \hline
    \end{tabular}

    \begin{tabular}{|l|c|}
        \multicolumn{2}{c}{Roll} \\
        \hline
        Model & mAP (\%) \\
        \hline
        Random & 41.44 $\pm 10.54$ \\
        3D Trajectory & 60.33 $\pm 5.92$ \\
        2D Trajectory & 60.56 $\pm 7.63$ \\
        2D Image & 70.33 $\pm 5.23$ \\
        \hline
    \end{tabular}
    \caption{Mean Average Precision (mAP) scores with 95\% confidence intervals for \textit{fall} and \textit{roll}, which exhibit significant performance difference for the 3D Trajectory and 2D image models. For all other verbs, there was no significant difference between models.}
    \label{tab:verbs}
\end{table}

Aside from these outlier cases, there are no significant differences in performance for each modality. These results suggest that, contrary to our hypothesis, 2D image-based models encode sufficient information to capture verb semantics on par with 3D trajectories-based models.

One possible explanation for these results is that 3D trajectories can be extracted from 2D inputs. To investigate whether this is the case, we perform a follow-up analysis using probing classifiers. Specifically, for each model, we fine-tune the encoder to predict the 3D position of the object at the final frame. We report the best Mean Squared Error (MSE) loss for each approach in Table \ref{tab:probe}. We see from this experiment that it is indeed the case that, from the 2D trajectory alone, the model is able to reconstruct the 3D trajectory reasonably well, and that adding the image to the 2D trajectory further improves the results. However, none of the 2D representations perfectly capture the 3D representation. Thus, one interpretation of this analysis in combination with the results from above is that, while 2D information is impoverished relative to 3D, the differences that are lost when moving from 3D to 2D are not differences that are central for differentiating verb semantics. 


\begin{table}[!htp]
    \centering
    \begin{tabular}{|l|c|c|}
        \hline
        Model & MSE \\
        \hline
        Random & 0.4104 \\
        3D Trajectory & 0.0104 \\
        2D Trajectory & 0.0409 \\
        2D Image & 0.0837 \\
        2D Image + 2D Trajectory & 0.0289 \\
        \hline
    \end{tabular}
    \caption{Mean Squared Error (MSE) for each model on a 3D object position regression task.}
    \label{tab:probe}
\end{table}

\section{Discussion \& Limitations}

Our results challenge the notion that richer environment representations will necessarily yield better representations of language, specifically in the context of verbs. Our experiments show that aside from narrow outlier cases discussed in Section \ref{sec:results}, models trained on 2D Image embeddings perform similarly well to models trained on 2D and 3D Trajectory data. Combined with followup probing analysis, our results suggest that, while some information is lost when collapsing to 2D world representations, the lost information might not be critical for differentiating verb semantics.

This work also highlights the challenges of language learning in embodied environments, and points toward the need to look deeper into how models may or may not be able to capture complex aspects of verb semantics, a major bottleneck in language understanding, which will be critical to building AI agents that interact effectively with humans in realistic environments.

A limitation of our work is that the data used in this study is a highly controlled environment with a relatively small number of verbs. This means that our results may not generalize to other datasets, or when scaling up to larger and more complex models. 
Thus, overall, these results do not close the book on the question of which world representations best support verb learning. 
Further research is needed to fully understand the potential and limitations of richer environment representations for language learning.

\bibliography{anthology,custom}

\newpage

\appendix

\section{Full Results}\label{app:full-results}

\begin{figure*}[!ht]
    \centering
    \includegraphics[width=\textwidth]{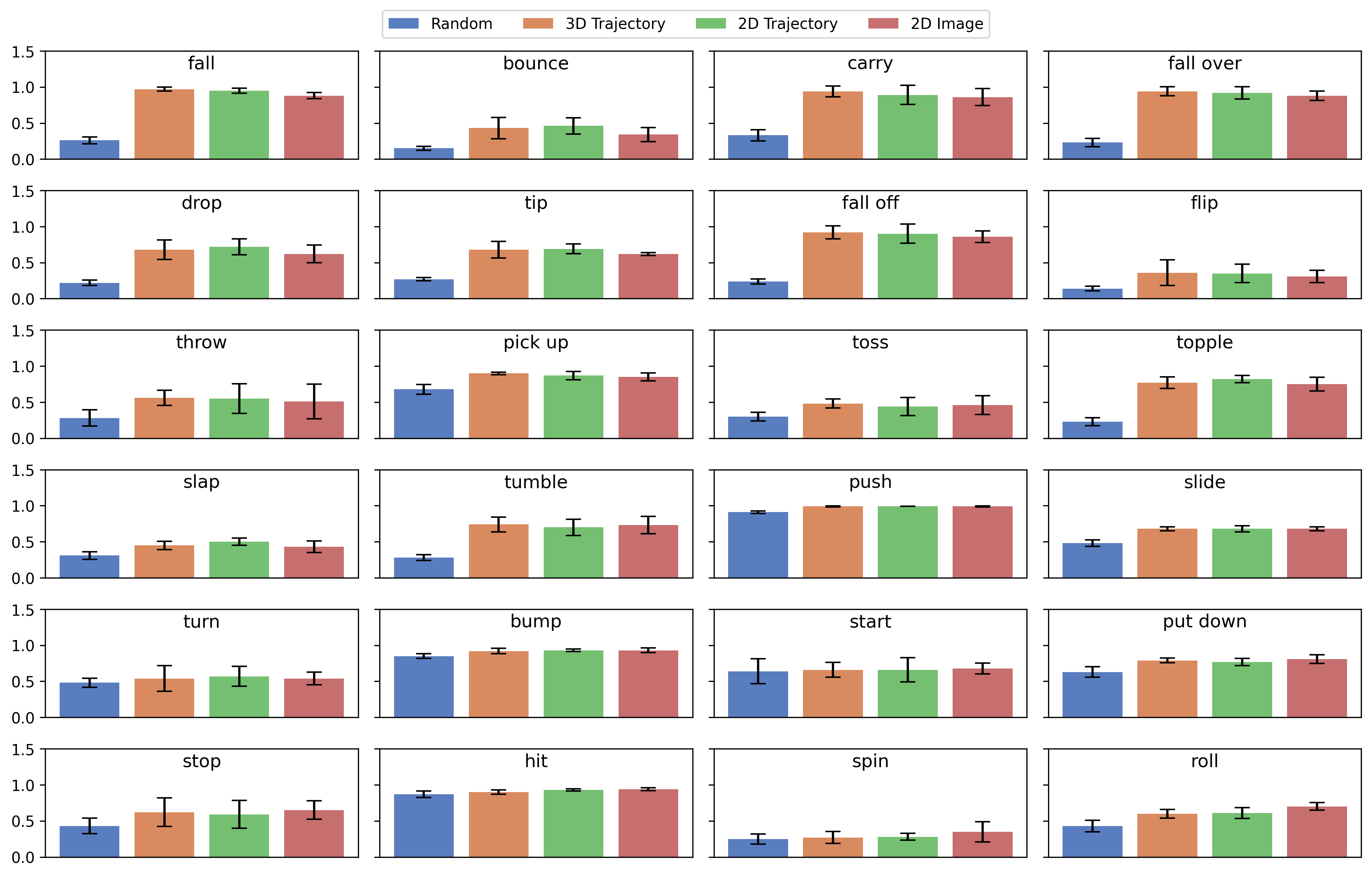}
    \caption{Results broken down by verb. Each subplot displays the mAP (macro) score of each approach on that verb.}
    \label{fig:results}
\end{figure*}

Figure \ref{fig:results} shows the Mean Average Precision broken down by verb.

\section{Convolution Experiments}\label{app:convolutional}

\begin{figure}[!ht]
    \centering
    \includegraphics[width=0.5\textwidth]{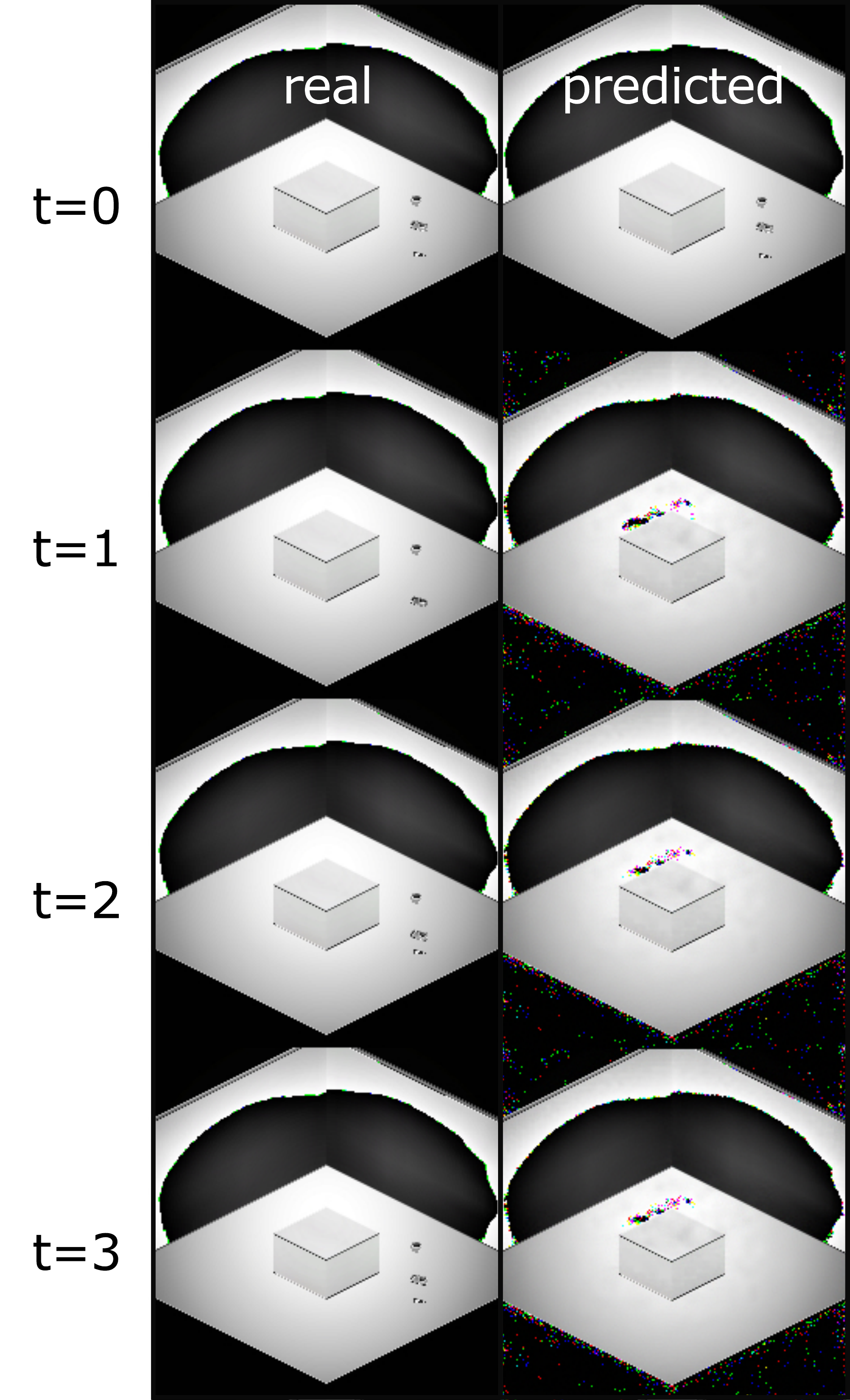}
    \caption{True and predicted frames during image-based pre-training. The frame at $t=0$ is the final input frame, where at $t=1..3$ are the real and predicted future frames. The predicted images suffer from mode collapse, and consistently reproduce the same reconstruction.}
    \label{fig:image-reconstructions}
\end{figure}

This section describes experiments run using a convolutional encoder on raw image data, rather than Inception-v3 embeddings. Figure \ref{fig:image-reconstructions} shows a side-by-side reconstructions that demonstrate the convolutional pre-training task suffering from mode collapse, where all predicted frames $t=1..3$ are the same, with the object somewhat focused around the center of the frame.

\end{document}